\documentclass[11pt,a4paper]{article}
\usepackage[hyperref]{acl2020}
\usepackage{times}
\usepackage{latexsym}
\usepackage{amsmath,amssymb,bm}
\usepackage{graphicx}
\usepackage{color,xcolor}
\usepackage{float}
\usepackage{subfigure}
\usepackage{booktabs}
\usepackage{mathtools}
\usepackage{multirow}
\usepackage{CJKutf8}

\usepackage{microtype}

\aclfinalcopy % Uncomment this line for the final submission
\def\aclpaperid{3352} %  Enter the acl Paper ID here

\title{DRTS Parsing with Structure-Aware Encoding and Decoding}

\author{
  Qiankun Fu$^{1,2,3}$, Yue Zhang$^{2,3}$, Jiangming Liu$^{4}$ \and Meishan Zhang$^{5}$ \\
  1. Zhejiang University \\
  2. School of Engineering, Westlake University \\
  3. Institute of Advanced Technology, Westlake Institute for Advanced Study \\
  4. ILCC, University of Edinburgh \\
  5. School of New Media and Communication, Tianjin University, China \\
  \texttt{\{fuqiankun,zhangyue\}@westlake.edu.cn} \and  \texttt{mason.zms@gmail.com} \\
  }

\date{}

\begin{document}
\begin{CJK}{UTF8}{gbsn}
\maketitle
\begin{abstract}
Discourse representation tree structure (DRTS) parsing is a novel semantic parsing task which has been concerned most recently.
State-of-the-art performance can be achieved by a neural sequence-to-sequence model, treating the tree construction as an incremental sequence generation problem.
Structural information such as input syntax and the intermediate skeleton of the partial output has been ignored in the model, which could be potentially useful for the DRTS parsing.
In this work, we propose a structural-aware model at both the encoder and decoder phase to integrate the structural information,
where graph attention network (GAT) is exploited for effectively modeling.
Experimental results on a benchmark dataset show that our proposed model is effective and can obtain the best performance in the literature.

\end{abstract}

\section{Introduction}
Discourse representation tree structure (DRTS) is a form of discourse structure based on Discourse Representation Theory  of \citet{Kamp:93}, a popular theory of meaning representation \citep{Kamp:81, Asher:93, Asher:03}.
It is designed to account for a variety of linguistic phenomena, including the interpretation of pronouns and temporal expressions within and across sentences.
Correspondingly, as one type of discourse parsing, DRTS parsing \cite{Liu:18} can be helpful for paragraph or document-level text understanding by converting DRS to tree-style DRTS. \cite{Liu:19}.

Figure 1 shows an example of DRTS, where the leaf nodes are discourse representation units (DRUs), upon which a discourse tree structure built.
In particular, a DRU consists of several individual tuples, where each tuple denotes a relation inside the DRU.
For example, there is a relationship ``\textit{That}" between the specific entity $x_{16}$ and a proposition $p_4$.
The relationships between the DRUs are organized by a tree skeleton,
which includes three types of nodes: the S(DRS) nodes to introduce DRU, the relation nodes for inter-DRU relationship, and the variable nodes, which are used to define S(DRS) (e.g., $p_4$, $k_1$ and $k_4$ ).

\begin{figure}[t]
\centering
\includegraphics[scale=0.6]{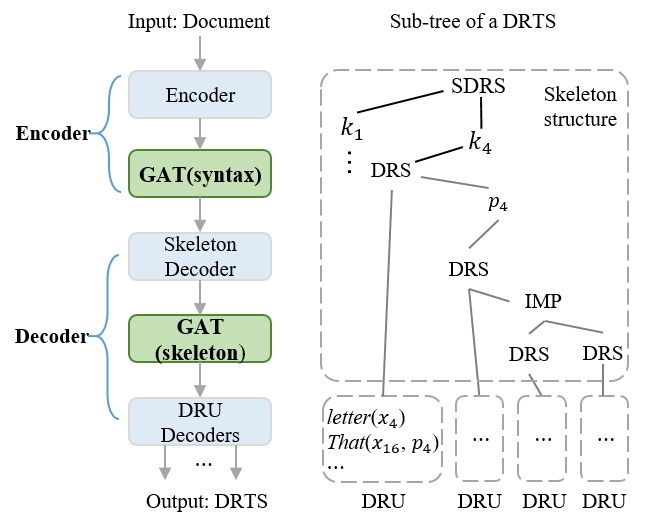}
\caption{Left: Our proposed model with two structure-aware module. Right: The DRTS for a clause in a document: ``The $\textcolor{purple}{\text{letter}}_{x_{4}}$ warns Jewish $\textcolor{purple}{\text{women}}_{x_{16}}$ \textit{that} $\colorbox{pink}{they will suffer if they date Arab men.}_{p_4}$"}
\label{fig:overall}
\end{figure}

There have been only a few existing studies related to DRTS parsing \citep{Noord:18-LREC, Noord:18-TACL}.
In particular, the end-to-end encoder-decoder model of \citet{Liu:19} gives the state-of-the-art performance, which converts the task into a sequence-to-sequence problem.
The input sequence consists of words in paragraphs, encoded by a BiLSTM structure,
and the output sequence is top-to-bottom depth-first traversal of the output DRTS tree,
which is decoded incrementally with an attention-based LSTM feature representation module.
During decoding, \citet{Liu:19} separate the skeleton generation and the DRU producing,
as illustrated by Figure \ref{fig:overall}.

Although highly effective, the above model ignores some useful structure information in both the encoder and the decoder,
which can be potentially useful for our task.
Specifically, for encoding, syntax-based tree structure information has been demonstrated effective for a number of NLP tasks \citep{Kasai:19, Li:18},
including several other types of discourse parsing \citep{Yu:18, Li:15}.
For decoding, the skeleton structure of DRTS can be also beneficial for our task.
As a two-phase decoding strategy is exploited,
the skeleton tree from the first phase could be helpful for DRU parsing of the second phase.

We propose to improve DRTS parsing by making use of the above structure information, modeling dependency-based syntax of the input sentences as well as the skeleton structure to enhance the baseline model of \citet{Liu:19} using Graph Attention Network (GAT) \cite{velickovic:18}, which has been demonstrated effective for tree/graph encoding \citep{Huang:19, Hu:19}.
In particular, we first derive dependency tree structures for each sentence in a paragraph from the Stanford Parser,
and then encode them directly via one GAT module, which are fed as inputs for decoding.
Second, after the first-state skeleton parsing is finished,
we encode the skeleton structures by another GAT module, feeding the outputs for DRU parsing.

Following \citet{Liu:19}, we conduct experiments on the Groningen Meaning Bank (GMB) dataset.
Results show that structural information is highly useful for our task, bring a significantly better performance over the baseline.
In particular, dependency syntax gives an improvement of 2.84\% based on the standard evaluation metrics and the skeleton structure information gives a further improvement of 1.41\%.
Finally, our model achieves 71.65\% F1-score for the task, 4.25\% better than the baseline model.
Additionally, our model is also effective for sentence-level DRTS parsing,
leading to an increase of 1.72\% by the F1-score by our final model.
We release our code and best models at \url{http://github.com/seanblank/DRTSparsing} for facilitating future research.

\begin{figure}[ht]
\centering
\includegraphics[scale=0.6]{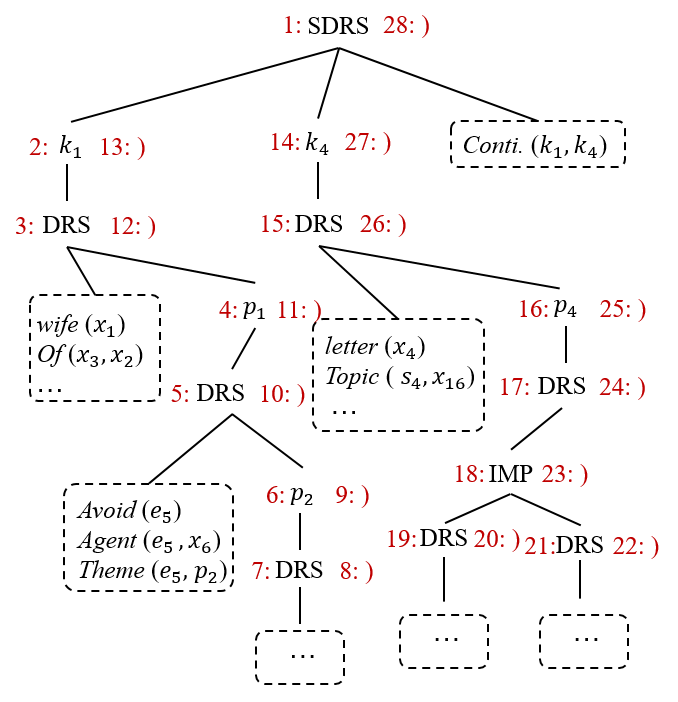}

\caption{A full DRTS tree for document: ``\textcolor{blue}{$k_1$:} At least 27 wives of Israeli rabbis have signed a letter urging Jewish women to avoid dating Arab men.
\textcolor{blue}{$k_4$:} The letter warns Jewish women that they will suffer if they date Arab men."
Red numbers indicate top-down depth-first order traversal of the DRTS skeleton.} \label{fig:DRTS}
\end{figure}
\section{Discourse Representation Tree (DRT) }\label{task:intro}

Formally, a DRT structure consists of two components according to the function:
(1) the leaf nodes  and (2) the tree skeleton (non-terminal nodes), respectively.
Similar to other types of discourse representation methods, we have minimum semantic units named by DRU,
and then a discourse tree is built by the discourse relationships between these minimum units.
Figure \ref{fig:DRTS} shows the full tree version of Figure 1 in the introduction.

\paragraph{DRU.}
DRU serves as terminal nodes of a DRT structure,
which is constituted by a set of unordered relation tuples,
as shown by the below dashed components of the tree in Figure \ref{fig:overall}.
A relation tuple consists of a relation $r$
and several arguments $v_1 \cdots v_n $ in $r$,
it can be denoted as $r(v_1 \cdots v_n)$.
Variables refer to entities $x$, events $e$, states $s$, time $t$, propositions $p$, segment $k$ and constants $c$.
The relation is used to indicate the discourse connections among the inside variables.
A total of 262 relation labels are defined in DRTS.
One DRU may include unlimited relation tuples,
which are all extracted from the corresponding text pieces.

\paragraph{Skeleton.}
The skeleton reflects the structural connection between DRUs.
Nodes in a skeleton can be divided into three categories,
including the (S)DRS nodes, the relation nodes and the variable nodes.
In particular, (S)DRS nodes denotes a full semantically-completed node of discourse analysis.
The relation node defines a specific discourse relationship over its covered (S)DRS nodes.
DRTS has defined six types of DRS relations,
including IMP (implication), OR (disjunction), DUP (duplex), POS (possibility), NEC (necessity) and NOT (negation), respectively,
which is orthogonal to the relations inside the DRUs.
The variable node assigns one (S)DRS node with a specific symbol.
There are two types of variable nodes, namely proposition and segment.
For example, in Figure \ref{fig:DRTS}, the root is a SDRS node, IMP is a relation nodes and $k_1$, $p_4$ denote the variable nodes.

\section{Baseline}
We take the multi-step encoder-decoder method of \citet{Liu:19} as the baseline model for DRTS parsing.
First, an encoder is used to convert one input paragraph into neural vectors by using word embeddings as well as BiLSTMs,
and then a multi-step decoder is exploited to generate a full tree structure in a sequential manner incrementally.

\subsection{Encoder}
Given a paragraph,
we concatenate all the sentences into one sequence, where each sentence is augmented with a start symbol $\langle s \rangle$ and an end token $\langle e \rangle$ at the front and end positions, respectively,
obtaining a final input sequence for the paragraph $D = \langle s \rangle,w_{1,1},...,w_{1,n_1},\langle e \rangle,\langle s \rangle,w_{2,1},..., w_{m,n_m},\langle e \rangle $.
For simplicity, we use $D = w_1,...,w_n$ to denote the sequence for short.

We use three different embedding representations to denote each word $w_i$:
\begin{equation}
    \bm{v}_i = \bm{e}_{\text{rand}}(w_i) \oplus \bm{e}_{\text{pret}}(w_i) \oplus \bm{e}_{\text{lem}}(w_i)  ,
\end{equation}
where $\bm{e}_{\text{rand}}(\cdot)$,
and  $\bm{e}_{\text{pret}}(\cdot)$ denotes random and pretrained embeddings for current word,
$\bm{e}_{\text{lem}}(\cdot)$ denotes the random embedding for current word lemma,
and $\oplus$ denotes concatenation,

We then apply MLP over the word representations,
and further use BiLSTM to encode the vector sequence:
\begin{equation}
\begin{split}
    \bm{x}_1 \cdots \bm{x}_n &= \text{MLP} (\bm{v}_1 \cdots \bm{v}_n) \\
    \bm{H}^{\text{enc}} = \bm{h}_{1} \cdots \bm{h}_n & = \text{BiLSTM}(\bm{x}_1 \cdots \bm{x}_n),
\end{split}
\end{equation}
where $\bm{H}^{\text{enc}} = \bm{h}_{1} \cdots \bm{h}_n$ is the encoder output.

\begin{figure*}[ht]

\centering
\includegraphics[scale=0.6]{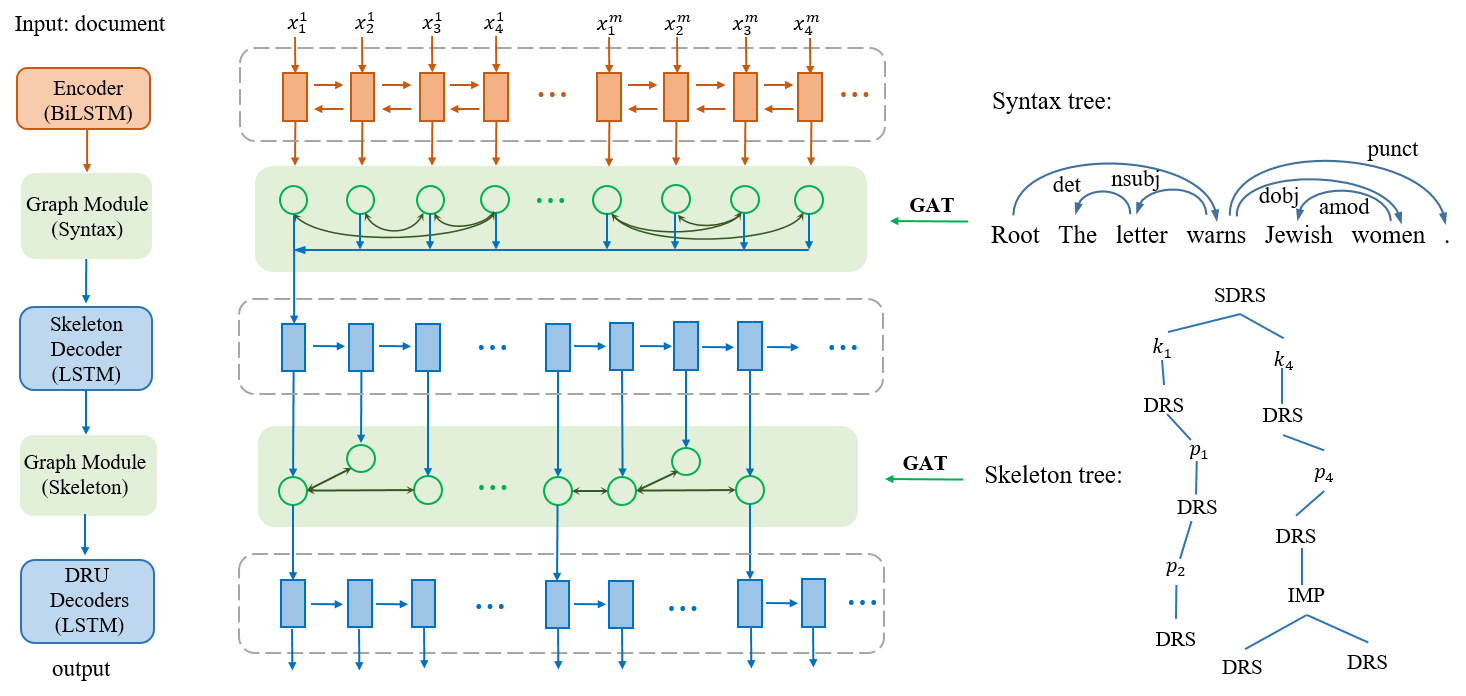}
\caption{Model structure with two graph modules over RNN outputs. The hidden state for each word of the encoder is taken as the input node vector of the GAT module using syntax structure, and the output is fed into the skeleton decoder.
The output of skeleton decoder is fed into the GAT module with the skeleton structure, and the output of GAT module is used to guide each DRU sequence generation.
}
\label{fig:model}
\end{figure*}

\subsection{Decoder}
We transform the DRTS structure into a sequence of symbols, so that the original DRTS can be restored from the symbol sequence as well.
By this transformation, we can apply the sequence-to-sequence architecture for decoding.
In particular, a two-stage strategy for the decoding is adapted,
first generating the skeleton structure,
and then generating the DRUs.
The key step is the transformation strategies of the two stages.

\paragraph{Generating the skeleton structure.}
We define two types of symbols for each skeleton,
where the first is the node label conjoined by a left bracket,
indicting the start of traversal of the current node,
and the second symbol is a right bracket,
indicting the end of traversal of the current node.
We exploit a top-down depth-first order to traverse the skeleton subtree, finishing a node traversal when all its child nodes have been finished.
Figure \ref{fig:DRTS} showed an example to illustrate the transformation.
In this way, we can obtain a symbol sequence  $Y^{\text{skt}}=y^{\text{skt}}_1, ..., y^{\text{skt}}_s$ which is equivalent to the skeleton tree.

\paragraph{Generating the DRUs.}
After the skeleton is ready, we start the DRU generation process.
The DRU nodes are only related to the (S)DRS nodes in the skeleton.
Thus we generate DRU nodes one by one according to the (S)DRS nodes in the skeleton structure.
For each DRU, we have two types of symbols, one for the relations and the other for the variables.
We first generate all the relations and then
generate the variables of each relation incrementally.\footnote{We follow a predefined order for relations. In fact, the order impacts little on the final influence. }
In this way,
we can obtain a sequence of $Y^{\text{dru}}=y^{\text{dru}}_1, ..., y^{\text{dru}}_t$
for DRU generation.\footnote{Our description is equivalent to \citet{Liu:19}, who split this process into two steps (i.e., relation prediction and variable prediction). We merge the relation and variable predictions for brief. }

\paragraph{Sequence decoding.}
We follow the standard sequence-to-sequence architecture \cite{Liu:18} to obtain the final sequence
$Y = Y^{\text{skt}} Y^{\text{dru}} = y^{\text{skt}}_1, ..., y^{\text{skt}}_s y^{\text{dru}}_1, ..., y^{\text{dru}}_t$ incrementally.
At each step, we score the candidate next-step symbols based on current observations:
\begin{equation}
    \begin{split}
      \bm{o}^{\text{skt}}_j  & = g^{\text{skt}}(\bm{H}_{y^{skt}_{<j}}, \bm{H}^{\text{enc}}), \\
      \bm{o}^{\text{dru}}_k  & = g^{\text{dru}}(\bm{H}_{y^{skt}_{<k}}, \bm{H}^{\text{skt}}, \bm{H}^{\text{enc}}),
    \end{split}
\end{equation}
where $\bm{H}^{\text{enc}}$ refers to the encoder outputs,
$\bm{H}^{\text{skt}}$ and $\bm{H}^{\text{dru}}$ denotes the outputs of skeleton decoder and the DRU decoder
uses left-to-right LSTMs over $Y^{\text{skt}}$ and $Y^{\text{dru}}$, respectively,
and $g^{\text{skt}}(\cdot)$ and $g^{\text{dru}}(\cdot)$ are neural feature extraction functions
for predicting skeleton and DRU symbols, respectively.
Here we neglect the detailed description for $g^{\text{skt}}(\cdot)$ and $g^{\text{dru}}(\cdot)$,
which can be found in \citet{Liu:19}.

\paragraph{Training.}
Given a set of labeled data, the model is trained to minimize average cross-entropy losses over all individual symbol predictions:
\begin{equation}
    L(\theta)=-\frac{1}{N} \sum_i \text{log} p_{y^{\text{*}}_i}
\end{equation}
where $\theta$ are the set of model parameters, $p_{y^{\text{*}}_i}$ denotes the output probability of $y^{\text{*}}_i$, which is computed by softmax over $\bm{o}^{\*}_i$,
$N$ is the total length of the output sequence.

\section{Structure-Aware Seq2Seq}
\label{sec:length}

To represent the structure features, we use a GAT module on top of encoder and skeleton decoder stage to enhance the baseline model.
The graph module is designed to learn non-local and non-sequential information from structural inputs.
In this section, we first describe the GAT in detail and then illustrate its application on our task.

\subsection{Graph Attention Network}\label{gat:intro}
Given a graph $G=(V, E)$, where each node $v_i$ has a initial vectorial representation,
the GNN module enriches node representation with neighbor informations derived from the graph structure:
\begin{equation}
    \bm{H}^{l+1} = \textsc{Gnn}(\bm{H}^{l}, \bm{A}; \bm{W}^{l}),
\end{equation}
where $\bm{H}^l \in \mathbb{R}^{n \times d}$ is the stacked hidden outputs for all nodes at layer $l$ ($\bm{H}^0$ denotes the input initial representations), $\bm{A} \in \mathbb{R}^{n \times n}$ denotes the graph adjacent matrix representation,
and $\bm{W}^l$ is the parameter set of the GNN at layer $l$.

Different information aggregation functions lead to different GNN architectures.
In particular, GAT uses the attention mechanism \cite{Bahdanau:14} on graph neighbors, which has been demonstrated more effective than graph convolution neural network (GCN).
The aggregation weights in GAT are computed by multi-head attention mechanism \cite{Vaswani:17}.

Specifically, given a node $i$ with a hidden representation $\bm{h}^{l}_i$ at layer $l$ and the its neighbors $\mathcal{N}_i$ as well as their hidden representations, a GAT updates the node's hidden representation at layer $l+1$ using multi-head attention:
\begin{equation}
        \bm{h}^{l+1}_i = \parallel_{k=1}^K \sigma (\sum_{j \in \mathcal{N}_i} \alpha_{ij}^k \bm{W}^k \bm{h}_j^l)
\end{equation}
where $\parallel$ represents concatenation, $\sigma$ is a sigmoid function, and $\bm{W}^k$ is the corresponding weight matrix of input linear transformation. $\alpha^k_{ij}$ are normalized attention coefficients computed by the $k$-th attention mechanism:
\begin{equation}
\begin{split}
    \alpha_{ij}^k &= \textsc{SoftMax}_j(e_{ij}) \\
    &= \frac{exp(e_{ij})}{\sum_{k \in \mathcal{N}_i} exp(e_{ik})}
\end{split}
\end{equation}
where $e_{ij}$ is attention coefficient that indicate the importance of node $j$ to node $i$ computed by:
\begin{equation}
    e_{ij} = \text{LeakyReLU}\big(f[\bm{W} \bm{h}_i \parallel \bm{W} \bm{h}_j]\big )
\end{equation}
$f(\cdot)$ is a single-layer feed-forward neural network, parameterized by a shared weight, $\bm{W}$ denotes a shared linear transformation and LeakyReLU is a non-linearity activation function.

\subsection{GAT for the Encoder}
On the encoder side, we equip the inputs with dependency syntax structures,
which have been demonstrated helpful for closely-related tasks such as RST discourse parsing.
A GAT module is used to represent the encoder output as mentioned in Section \ref{gat:intro}.
We transform the document into a dependency graph represented by a undirected adjacent matrix using an off-the shelf dependency parser \cite{Chen:14}.
The hidden states of each node is updated with a multi-layer GAT network on the adjacent matrix $\bm{A}$:
\begin{equation}
    \bm{H}^{\text{g-enc}} = \textsc{Gat}^{\text{enc}}(\bm{H}^{\text{enc}}\oplus \bm{E}^{\text{syn}}, \bm{A}; \bm{W}),
\end{equation}
where $\bm{E}^{\text{syn}}$ is the embedding outputs of the syntactic labels in the dependency tree.

The learned representation $\bm{H}^{\text{g-enc}}$ is used to substitute the original $\bm{H}^{\text{enc}}$ for predictions.

\subsection{GAT for the Decoder}
We further enhance the baseline model by exploiting the partial output after skeleton prediction step is finished.
On one hand, the skeleton structures can guide for DRU parsing.
On the other hand, the joint skeleton and DRU parsing can further help to rerank the skeleton predictions as well,
since global skeleton representations are exploited.

Specifically, after all the skeleton nodes are generated, we construct a graph based on the nodes except the right parenthesis as shown in Figure \ref{fig:model}.
We use a GAT network on top of the hidden states to capture global structure information:
\begin{equation}
    \bm{H}^{\text{g-skt}} = \textsc{Gat}^{\text{skt}}(\bm{H}^{\text{skt}}\oplus \bm{E}^{\text{skt}}, \bm{A}; \bm{W}),
\end{equation}
where $\bm{E}^{\text{skt}}$ is the embedding outputs of the node labels in the generated skeleton tree,
and the global skeleton-aware representation $\bm{H}^{\text{g-skt}}$ is used instead of the original $\bm{H}^{\text{skt}}$ for future predictions.

\begin{table}[t]
\centering
\begin{tabular}{lcccc}
\hline
\textbf{Section} & \textbf{\#Doc} & \textbf{\#Sent}  &$\textbf{AVG}_{\text{sent}}$ &$\textbf{AVG}_{\text{word}}$      \\
\hline
Train & 7843  & 48599 & 6.2 & 135.3 \\
Devel   & 991  & 6111   & 6.2 & 134.0 \\
Test  & 1035  & 6469  & 6.3 & 137.2 \\
\hline
\end{tabular}
\caption{\label{tab:dataset} Statistics on GMB document level benchmarks,  $\textbf{AVG}_{\text{sent}}$ and $\textbf{AVG}_{\text{word}}$ denote the average number of sentences and words per document, respectively. }
\end{table}

\section{Experiments}
\subsection{Data and Settings}
\paragraph{Data}
We conduct experiments on the benchmark GMB dataset,
which provides a large collection of English texts annotated with Discourse Representation Structures \cite{Bos:17}.
We follow \citet{Liu:19} using the processed tree-based DRTS format, and focus on document-level parsing.
The data statistics are shown in Table \ref{tab:dataset}.

\paragraph{Hyperparameters}
We exploit the same hyper-parameters as \citet{Liu:19} for fair comparison.
In particular, we use the same pre-trained 100-dimensional word embeddings, which are trained on the AFP portion of the English Gigaword corpus.
The sizes of random word and lemma embeddings are set to 300 and 100, respectively.
The hidden sizes of BiLSTM modules in encoder and decoder are set to 300 and 600, respectively.
In addition, the BiLSTM layer sizes of encoder and decoder are respectively 2 and 1.
The hidden size of GAT modules is set to 300 and 600 for encoder and decoder, respectively.

\subsection{Evaluation}\label{eval}
Following \newcite{Liu:19}, we adopt the \textsc{Counter} \cite{Noord:18-LREC} tool to evaluate our final experimental results.
In particular, we first transform the DRTS into a clause format and then run the standard evaluation script to obtain the F1-scores of our results compared with the gold-standard clause form.
Note that \textsc{Counter} is computationally expensive, requiring more than 50 hours for the entire test dataset by using more than 100 threads.
To facilitate development and analysis experiments,
we suggest three alternatives for evaluation particularly for development experiments:
\begin{itemize}
  \item[(1)]BLEU: a standard BLEU \cite{Papineni:02} value is adopted as the metric to evaluate the resulting node sequence against the gold-standard output,
  since we model the task as a sequence-to-sequence task.
  \item[(2)]Skeleton: The bracket scoring method of constituent parsing is exploited to evaluate the skeleton performance,
  by regarding terminal DRU nodes as words in comparison with a constituent tree.\footnote{https://nlp.cs.nyu.edu/evalb/}
  \item[(3)]Tuple: The F1-score of tuple-level matching is exploited to measure the DRU performance, since the basic units inside a DRU are tuples of relation-variable functions.
   Exact matching is adopted considering variable orders.
\end{itemize}
The BLEU is used for development and the Skeleton and Tuple are used for analysis.

\subsection{Development Experiments}

We conduct experiments on the development dataset to understand the key factors of our proposed model.

\begin{figure}[tb]
\begin{center}
	\subfigure[structure labels]{\label{fig:label:emb}
		\centering{\includegraphics[scale=1.2]{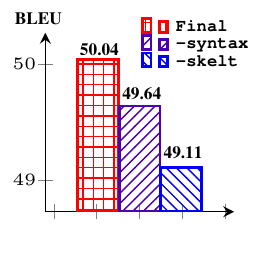}}
	}
\hspace{-0.3cm}
	\subfigure[GAT modules]{\label{fig:gat:module}
		\centering{\includegraphics[scale=1.2]{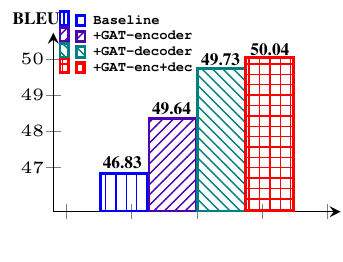}}
	}
\caption{Feature ablation experiments. }\label{dev:bar}
\end{center}
\end{figure}

\paragraph{Impact of structure labels}
Syntactic arcs and skeleton labels are embedded and concatenated to the embedding of the current node
when using GAT to model the tree structure.
We conduct a comparison to examine their effectiveness in our model.
Figure \ref{fig:label:emb} shows the results.
We can see that a performance degradation occurs without these label embeddings.
In particular, BLEU score drops by 0.4 without syntax label embeddings and 0.93 without skeleton label embeddings,
which shows that modeling label information improves unfixed skeleton tree structure even more.

\paragraph{Impact of GAT setting}

As our proposed modules involve a $l$-layer GAT, we investigate the effect of the layer number $l$ on the dev set as shown in Table \ref{tab:gat}.
In particular, we vary the value of $l$ in the set \{1, 2, 3, 4, 5\} and measure the corresponding BLEU scores.
The structural-aware model equipped with GAT achieves the best performance when $l$ is 2, which justifies the selection on the number of layers in the experimental setting section.
Moreover, a dropping trend on both metrics is present as $l$ increases.
For a larger $l$, the GAT module becomes more difficult to train due to larger amounts of parameters.
One intuitive reason is that each layer of the GAT module aggregates the direct neighbor information of a node.
After 2 layers, each node can obtain sufficient information, and further more layers can bring noise.

\begin{table}[t]
\centering
\begin{tabular}{lcc}
\hline
\textbf{Model}  & \textbf{BLEU} \\
\hline
Head=1           & 48.76 \\
Head=2           & 49.48 \\
Head=3           & 50.01 \\
Head=4           & \bf 50.04 \\
Head=5           & \bf 50.04 \\
\hline
\end{tabular}
\begin{tabular}{lcc}
\hline
\textbf{Model}  & \textbf{BLEU} \\
\hline
layers=1         & 49.11 \\
layers=2         & \bf 50.04 \\
layers=3         & 49.72 \\
layers=4         & 49.01 \\
layers=5         & 48.54 \\
\hline
\end{tabular}
\caption{\label{tab:gat} GAT settings results on development set. }
\end{table}

We make comparison with multi-head attention, varying the heads in the set \{1, 2, 3, 4, 5\} and checking the corresponding BLEU scores.
Theoretically, the larger the number of heads, the better the performance of the model.
As can be seen in Table \ref{tab:gat}, when the number of heads exceeds 4,
the performance becomes relatively stable. We thus choose the head to be 4 for the remaining experiments.

\paragraph{Influence of the encoder and decoder GAT modules}
As shown in Figure \ref{fig:gat:module}, without using structure information, the baseline encoder-decoder \cite{Liu:19} model gives a development BLEU of 46.83.
Adding a GAT module to the encoder as described in Section 4.2 increases the BLEU score to 48.35, demonstrating the usefulness of syntax-aware module.
Furthermore, adding a GAT module to the decoder as described in Section 4.3 improves the performance to 49.73, which shows that our skeleton structure model is useful.
Finally, a combination of both gives a 50.04 BLEU score.

\subsection{Final Results}

\begin{table}[t]
\centering
\begin{tabular}{c|cc}
\hline
\textbf{Model}  & \textbf{BLEU}  &\textbf{exact F1}\\
\hline
\newcite{Liu:19}        &  \multirow{2}{*}{46.86}     & \multirow{2}{*}{66.56} \\
(baseline)          &       &   \\
\hline
GAT-encoder         & 48.24     & 69.40 \\
GAT-decoder         & 50.04     & 70.81 \\
GAT-enc+dec         & \bf 50.16     & \bf 71.65 \\
\hline\hline
Tree-LSTM           & 48.36     & 69.66 \\
GCN                 & 49.88     & 70.72 \\
\hline
\end{tabular}
\caption{\label{tab:test} Final results on the test dataset. }
\end{table}

\begin{table}[t]
\centering
\begin{tabular}{c|cc}
\hline
\textbf{Model}  & \textbf{BLEU}  &\textbf{exact F1}\\
\hline
\newcite{Liu:19}         & \multirow{2}{*}{64.96}     & \multirow{2}{*}{77.85} \\
(baseline)          &       &   \\ \hline
GAT-encoder              & 66.02     &  78.22 \\
GAT-decoder              & 66.69     &  79.14 \\
GAT-enc+dec              & 68.14     &  79.94 \\ \hline
\newcite{Liu:18}         & 57.61     & 68.72 \\
\hline
\end{tabular}
\caption{\label{tab:sent} Results on the sentence-level dataset. }
\end{table}

Table \ref{tab:test} shows the final results on the GMB test dataset.
We report performances of the baseline and various tree-structure systems using the exact F1-score by \textsc{Counter} in addition to BLEU.
The observations are consistent with the development set.
Our final model, the joint GAT-enc+dec model, achieves competitive performance, with a exact F1-score of 71.65\%.
Our GAT enhanced models outperform the state-of-the-art model.
For the vanilla encoder-decoder model, our GAT-encoder obtains a absolute improvement of 2.84\% exact F1-score,
which demonstrates that modeling syntax information is useful.
The GAT decoder improves the performance to 70.81\%, giving a 4.25\% promotion, which indicates that the skeleton structure is helpful to DRTS parsing.

As shown in Table \ref{tab:test},
Tree-LSTM and GCN based systems also give competitive results to the state-of-the-art baseline model,
which again demonstrates the effectiveness of modeling tree structures.
GCN achieves better performance than Tree-LSTM by 1.06\%, which can be because the GNN-based model obtains global information during layer stacking, but Tree-LSTM can only capture local structural information.
GAT performs better than GCN by 0.84\%, showing that GAT is a competitive choice of GNN.
Consistent with observations of BLEU scores, our proposed GAT-enc+dec model shows the best performance on both evaluation metrics.

In addition, we perform experiments on sentence-level datasets as shown in Table \ref{tab:sent} as well, following \newcite{Liu:19}.
We use the same setup as the document-level structure-aware model.
As shown, both the GAT encoder and decoder can bring better results (i.e., 0.37\% and 1.29\% by the GAT encoder and decoder, respectively),
and their combination can give further improvements (i.e., 0.80\% over the GAT-decoder) significantly,
which are consistent with the findings of the document-level parsing.
Finally, the sentence-level performance reaches 79.94\%, a new state-of-the-art score.
The results demonstrate that our model is also applicable to sentence-level DRTS parsing.

\begin{figure}[t]
\centering
\includegraphics[scale=1.1]{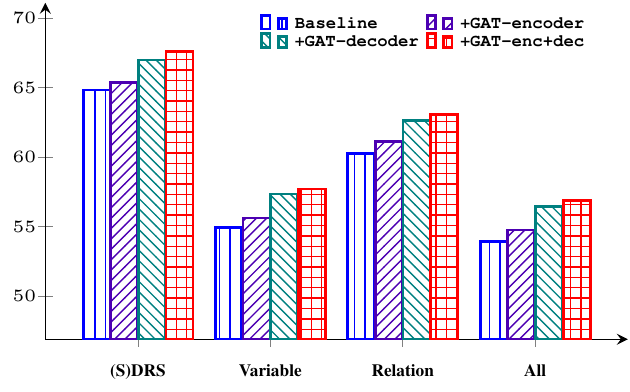}
\caption{ Skeleton-level evaluation F1 (\%) results.}
\label{fig:skt}
\end{figure}

\setlength{\tabcolsep}{3.5pt}
\begin{table}[t]
\centering
\begin{tabular}{lcccc}
\toprule
\textbf{Model}  & $\text{Rel}_{-\text{var}}$   & Rel    & Unary  & Binary   \\
\hline
baseline         & 64.13     & 34.80   & 39.21 & 26.38   \\
GAT-encoder      & 66.67     & 36.08   & 40.98 & 27.10  \\
GAT-decoder      & 68.32    & 36.41   & 42.34 & 27.22   \\
GAT-enc+dec      & \bf 68.97    & \bf 37.09   & \bf 43.76 & \bf 27.74\\
\bottomrule
\end{tabular}
\caption{\label{tab:rel} Relation-level evaluation F1 (\%) results. }
\end{table}

Interestingly, we find that the BLEU metric is highly indicative of model performance.
Based on the observed pair of values on the test results, we are able to approach the correction between BLEU and COUNTER by a line appropriately,
demonstrating a faithful alignment to the COUNTER metric.
The observation indicates that the BLEU is also a good metric for the task.
Noticeably, one advantage of the BLEU is that the metric calculation is much faster (i.e., only several seconds) than the exact-F1 score,
since the latter one consumes at least 24 hours as well as 100G+ memory for the evaluation of the test dataset.

\subsection{Analysis}
We conduct analysis to examine benefits by the structural-aware model.
As the decoding process is decomposed into two steps,
we examine the respective gains with respect to the two components, namely skeleton prediction and DRU parsing.

\paragraph{Influence on Skeleton Prediction}

\begin{figure*}[ht]

\centering
\includegraphics[scale=0.7]{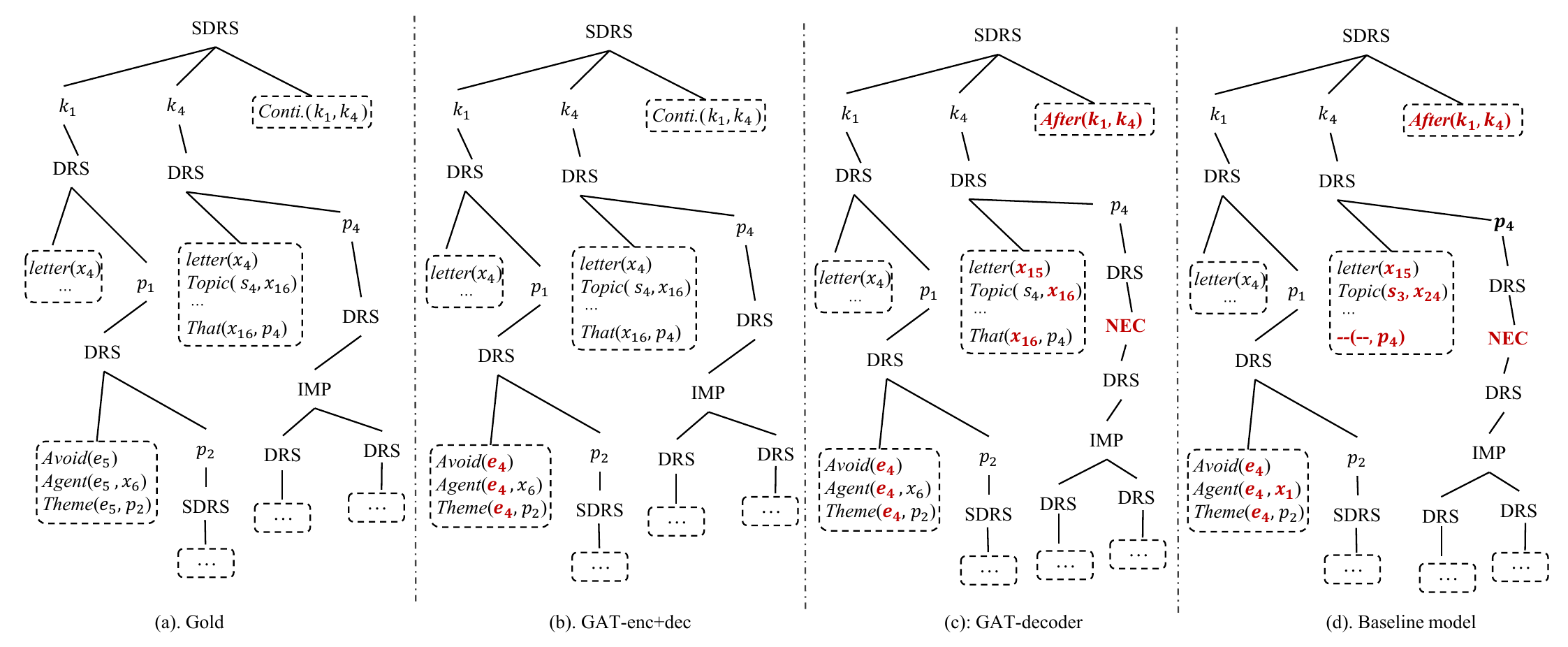}
\caption{Discourse representation tree structure examples generated by different models: ``\textcolor{blue}{$k_1$:} At least 27 wives of Israeli rabbis have signed a letter urging Jewish women to avoid dating Arab men.
\textcolor{blue}{$k_4$:} The letter warns Jewish women that they will suffer \textcolor{red}{\bf if} they date Arab men."
}
\label{fig:case}
\end{figure*}
The bracket scoring metric suggested in Section \ref{eval} is used to measure the performance of skeleton prediction.
Figure \ref{fig:skt} shows the F1-scores with respect to node types,
which are categorized into three types (Section \ref{task:intro}), namely (S)DRS, relation and variable.
In addition, the overall performance is reported as well.
First, we can see that the (S)DRS nodes can achieve the best performance across the three types,
the relation nodes rank the second and the variable type has the worst performance.
This indicates the relative difficulty in parsing the three types of nodes.
In particular, locating a DRU is relatively simpler as (S)DRS connects with DRU directly,
followed by the coarse-grained discourse relations over the DRUs,
while variable nodes are much more difficult since the order matters much (i.e., the subscript number in the variable).
Second, the tendencies in terms of different models on the three categories are the same as the overall tendency,
where our final model can bring the best skeleton performance,
and the baseline shows the worst performance.
The observation demonstrates the robustness of our proposed structural-aware model: we can achieve consistently better performances on all the types over the baseline.

\paragraph{Influence on relation tuples inside DRUs}

Further we analyze the model performance on DRU parsing.
A strict matching strategy on the relation tuples inside DRUs is used to measure the performance,
as described in Section \ref{eval}.
Table \ref{tab:rel} shows the performances,
where the F1-scores of the overall matching, only relation matching as well as unary and binary relation tuples are reported.\footnote{There are no relations containing more than two variables according to the corpus statistics.}
First, we can find that the overall exact matching F1-score is rather low (below 40).
When considering the relation performance ignoring the variables, the final F1-score reaches, with an increase of 31.88,
which indicates that variable recognition is extremely difficult.
Variables in DURs are similar to the variable nodes in skeleton,
however the scale of the inside DRU variables is much larger.
We further categorize the relation tuples by their number of variables.
The unary tuples (i.e. tuples consist of only one variable node) can obtain better performance than the binary tuples (i.e. tuples consist of two variable nodes), which is reasonable.
In addition, we look into the performance in terms of different models.
We can see that all structural-aware models can obtain better performances than the baseline on all settings,
demonstrating the effectiveness of our proposed models.
In particular, the GAT-decoder demonstrates relatively higher performance compared to GAT-encoder, which is consistent with the results observed in Table \ref{tab:test}.
As expected, the final joint GAT-enc+dec model obtain a better score than both of individual GAT models.

\paragraph{Case study}
Figure \ref{fig:case} shows one case study to illustrate the gains of our proposed models over the baseline model,
where the detailed differences are highlighted with red color.
As shown, the baseline model is already able to obtain a strong results with linguistically-motivated copy strategies, constraint-based inference and so on.
However, without structural-aware information,
the model is ineffective to handle several implicit long-distance dependencies.

For example, the relation of ``That($x_{16}, p_4$)'' is unable to be recognized by the baseline model,
while the models with structural-aware GAT decoder can get it correctly.
The major reason is that the structural-aware decoder can transmit the information from $p_4$ to its parent node,
which can facilitate the next-step generation of the parent node.

On the other hand, the syntactic information from the input sentences can help the first-step skeleton disambiguation.
For example, as shown in Figure \ref{fig:case}, the models without GAT-encoder can misclassify the relations between $k_1$ and $k_4$,
which is the discourse relation between the input two short sentences.
The major reason of the misleading may be possibly due to the word ``if'' in the second sentence,
which is one indicator for the \texttt{After} relation.
When the syntactic information is encoded by the GAT encoder,
the GAT-enc+dec model can learn the fined-grained dependency reduced by the word ``if'',
and thus is able to obtain the accurate relation of the two sentences (i.e., \texttt{Conti.})

\section{Related work}
Discourse parsing is one important topic in the NLP.
There are several main types of discourse parsing tasks in the literature, including rhetorical structure theory (RST; \citealp{RST}) based parsing,
centering theory (CT; \citealp{Grosz:95,Barzilay:08}) based parsing and DRT based parsing in this study.

Discourse Representation Theory (DRT) based parsing is a relatively classic, yet not fully researched semantic analysis task because of its complexity.
\citet{Le:12} present the first work of a data-driven DRT parser, using a graph-based representation of DRT structures.
Recently, \citet{Noord:18-TACL} apply the idea of neural machine translation for graph-based DRT parsing, achieving impressing performance.
These studies only focus on sentence-level DRT representations, as the complexity would increase much at the paragraph level.
In contrast, we investigate the paragraph level DRT parsing.

DRTS parsing simplifies graphs into trees.
There are two existing papers in this line.
\citet{Liu:18} are the first to work on DRTS parsing,
who propose an end-to-end sequence-to-sequence model for the task.
Further, \citet{Liu:19} improve the model by suggesting several effective strategies including supervised attention, copying from alignments, and constraint-based inference.
In this work, we improve DRTS parsing instead of \citet{Liu:19} with two types of structure information.

Syntax information has been widely exploited for NLP tasks.
Seminal work exploits discrete features designed by experts \citep{Feng:14,Heilman:15}.
Recently, a range of neural modules have been proposed to encode syntax, such as Tree-LSTM \citep{Tai:15,Zhu:2015,Teng:16}, Tree-CNN \cite{Roy:20} and
the recently proposed implicit approaches \citep{Yin:18, Zhang:19}.
Syntax has been demonstrated effective for RST based discourse parsing as well \cite{Yu:18}.
Our work is to build a syntax tree-aware model and we are the first to use syntax for DRT based discourse parsing.

GNN has received increasing interests for its strong capability
of encoding structural information \citep{Kipf:16,Bastings:17,Zhang:18-S-LSTM,Zhang:19-SA,Song:18}.
GAT is one representative model,
which demonstrates success in a number of NLP tasks \citep{Huang:19,Hu:19}.
In this work,
we exploit GAT to represent tree-structural information for DRTS parsing.

\section{Conclusion}
We investigated the representation of structural information for discourse representation tree structure parsing, showing that a graph neural network can bring significant improvements. In particular, we use GAT for representing syntax in encoding, and representing a structural backbone for decoding. Experiments on the standard GMB dataset show that our method is high effective, achieving the best results in the literature.

\section*{Acknowledgments}
We thank all reviewers for the valuable comments, which greatly help to improve the paper.
This work is supported by the National Natural Science Foundation of China (NSFC No. 61976180),
the funds of Beijing Advanced Innovation Center for Language Resources (No. TYZ19005)
and the Westlake University and Bright Dream Joint Institute for
Intelligent Robotics. Meishan Zhang is the corresponding author.

\bibliography{acl2020}
\bibliographystyle{acl_natbib}
\end{CJK}
\end{document}